\documentclass[10pt,twocolumn,letterpaper]{article}

\usepackage{mmstyle}
\usepackage{iccv}
\usepackage{times}
\usepackage{epsfig}
\usepackage{graphicx}
\usepackage{amsmath}
\usepackage{amssymb}
\usepackage{color}
\usepackage{algorithm}
\usepackage{algorithmicx}
\usepackage{subfigure}
\usepackage{algpseudocode}
\usepackage{booktabs}  
\usepackage{threeparttable}  
\usepackage{multicol}  
\usepackage{multirow}  
\usepackage{tabu}  
\usepackage{float}
\usepackage{caption}

\usepackage[pagebackref=true,breaklinks=true,letterpaper=true,colorlinks,bookmarks=false]{hyperref}

 \iccvfinalcopy % *** Uncomment this line for the final submission
 
\def\iccvPaperID{3327} % *** Enter the ICCV Paper ID here

\ificcvfinal\fi

\begin{document}

%%%%%%%%% TITLE
\title{Memory-Based Neighbourhood Embedding for Visual Recognition}

%% omit the following lines up until the closing ``}''.
%% Additional authors and addresses can be added with ``\and'',
%% just like the second author.
%% To save space, use either the email address or home page, not both
%\and

\author{Suichan Li$^{1,2}$\thanks{This work is done when Suichan Li is an intern at SenseTime.}\\
\and
Dapeng Chen$^{3}$\thanks{D. Chen and B. Liu are the co-corresponding authors.}\\
\and 
Bin Liu$^{1,2}$\footnotemark[2]\\
\and
Nenghai Yu$^{1,2}$\\
\and
Rui Zhao$^3$\\
\and
$^1$School of Information Science and Technology, University of Science and Technology of China\\
$^2$Key Laboratory of Electromagnetic Space Information, the Chinese Academy of Sciences\\
$^3$SenseTime Research\\
{\tt\small lsc1230@mail.ustc.edu.cn, \{chendapeng,zhaorui\}@sensetime.com, \{flowice,ynh\}@ustc.edu.cn}
%{\tt\small \{flowice,ynh\}@ustc.edu.cn}
}

\maketitle
\thispagestyle{empty}

%%%%%%%%% ABSTRACT
\begin{abstract}
Learning discriminative image feature embeddings is of great importance to visual recognition. To achieve better feature embeddings, most current methods focus on designing different network structures or loss functions, and the estimated feature embeddings are usually only related to the input images. In this paper, we propose Memory-based Neighbourhood Embedding (MNE) to enhance a general CNN feature by considering its neighbourhood. The method aims to solve two critical problems, i.e., how to acquire more relevant neighbours in the network training and how to aggregate the neighbourhood information for a more discriminative embedding. We first augment an episodic memory module into the network, which can provide more relevant neighbours for both training and testing. Then the neighbours are organized in a tree graph with the target instance as the root node. The neighbourhood information is gradually aggregated to the root node in a bottom-up manner, and aggregation weights are supervised by the class relationships between the nodes. We apply MNE on image search and few shot learning tasks. Extensive ablation studies demonstrate the effectiveness of each component, and our method significantly outperforms the state-of-the-art approaches.

\end{abstract}

%%%%%%%%% BODY TEXT

\section{Introduction}

Encoding the semantic information of an image into a feature embedding is a core requirement for visual recognition. Images from the same or related classes are desired to be mapped to nearby points on a manifold, which is critical to many applications like few-shot learning~\cite{vinyals2016matching, snell2017prototypical, sung2018learning}, visual search~\cite{hadi2015buy, li2014deepreid, weinberger2006distance}, face/person recognition~\cite{parkhi2015deep, schroff2015facenet, wang2017normface, Chen_2018_CVPR} and fine-grained retrieval~\cite{oh2016deep, Song_2017_CVPR}. With ideal feature embeddings, classification tasks could be reduced to the nearest neighbour problem, while retrieval tasks would be made easier by examining the inter-image relationships.

\begin{figure}[t]
	\centering
	\includegraphics[width=0.98\linewidth]{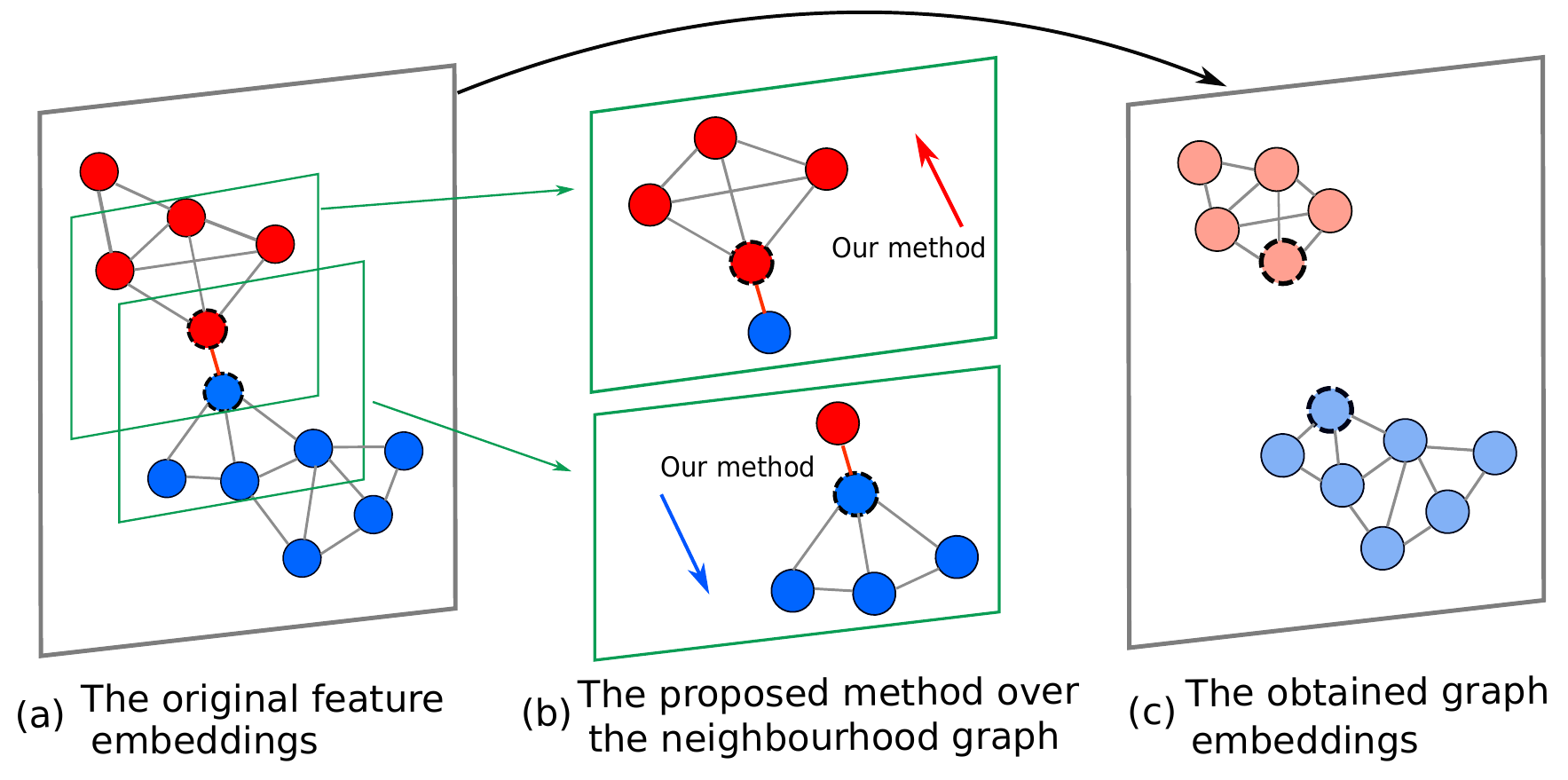}
	\caption{\small
  A case to explain the basic idea of the proposed method. The red nodes and blue nodes represent the feature embeddings of two classes respectively. The original feature embeddings may not be able to separate the samples of different classes as the two nodes connected by the red line. While our approach, taking the feature and its neighbourhood graph into account, can modify the current feature embeddings with their context structure and output the embeddings more consistent with their labels. } \label{fig:case}
 \vspace{-1em}
\end{figure}

To improve the feature embeddings with deep neural networks, state-of-the-art methods mainly focused on designing network structures. For example, VGGNets~\cite{simonyan2015very} and GoogleNet~\cite{szegedy2015going} suggested increasing the depth of a network can improve the quality of feature embedding, while ResNet~\cite{he2016deep} showed that adding the identity-based skip connections can help to learn deeper and stronger networks. 
%Recently, researchers have  proposed to automatically search better network architectures by reinforcement learning~\cite{baker2017designing}, evolutionary algorithm~\cite{real2017large} as well as gradient-based method~\cite{liu2018darts}.
At the same time, there also exists a stream of methods which supervise the feature embeddings by imposing different loss functions, including the contrastive loss~\cite{hadsell2006dimensionality}, the triplet loss~\cite{weinberger2006distance} and quadruplet loss~\cite{chen2017beyond}. Chen \emph{et al.}~\cite{Chen_2018_CVPR} built a graph within an image batch, forming a kind of group loss with CRF. The main goals of these loss functions are to reduce the intra-class variations while enlarging the inter-class variations. 

Although these efforts continuously bring better feature embeddings, most of them estimate a feature embedding based on a single image, and the abundant contextual information is ignored. When the same category individuals undergo drastic appearance changes, or visual differences between different categories are small,
%which is easily overwhelmed by factors such as pose, viewpoint, or location of the object in the image, 
 it becomes very difficult for the embedding of a single image to separate the samples of different classes. The example of such a situation is illustrated in Fig. \ref{fig:case}a, where the feature embeddings of the two classes can be very close to each other. Nevertheless, if taking the feature embedding and its neighbourhood into account,  two close feature embeddings can be classified two different clusters, and the affinities to their neighbours can be in turn used to modify original feature embeddings, achieving more discriminative features as demonstrated in Fig. \ref{fig:case}b and Fig. \ref{fig:case}c.

To effectively enhance a feature embedding and to take full advantage of the identity annotations, we propose the Memory-based Neighbourhood Embedding (MNE) to exploit abundant features and relationships among the neighbours.  For a conventional neural network, the batch-based training strategy determines that a training sample can only observe the features of images in the same batch. To acquire more relevant contextual information, we augment the network with an episodic memory, which stores the necessary features and labels of many image instances, and the memory is updated according to the newly computed CNN features during training. For every image in the image batch, we retrieve the neighbours from memory and organize them to be a tree graph, where the target instance corresponds to the root node. Based on the tree structure, we propose an iterative aggregation strategy to gradually propagate the neighbours' information to the root node.  In this way, neighbours close to the target instance can have a more significant influence over the embedding of target instance than the remote ones. We also have observed that if two nodes belong to the same class, a larger aggregation weight between them is preferred, which can help to produce more compact neighbourhood embeddings within each class. A new attention module is therefore introduced to predict the aggregation weights,  and we supervise the module by the node-wise class relationships. After iterations of feature aggregation, we obtain the neighbourhood embedding of the target instance from the root node.

The contributions of this paper could be summarized into three-fold: (1) We exploit the neighbourhood information for feature embedding by augmenting the episodic memory to the deep neural network. The memory can provide more relevant neighbours and support end-to-end training. (2) An iterative feature aggregation strategy is proposed to summarize the information in the neighbourhood. The strategy organizes the neighbourhood as a tree graph and gradually propagate the relevant information to the root node. (3) A new attention module is introduced for the feature aggregation, which is additionally supervised by the node-wise semantic relationships to better separate the feature embeddings of different classes. We apply MNE on image search and few shot learning tasks. Extensive ablation studies validate the effectiveness of the three components, and our method significantly outperforms the state-of-the-art approaches.  

\section{Related Work}

The proposed method aims to exploit the context for feature learning.
It is related to Graph Neural Networks (GNN) that aggregate information from graph structure, and share some similarities with the methods in transductive few-shot learning.

\noindent\textbf{Context-based Feature Learning}. Recently, researchers started to exploit the context information with the deep neural network to
learn better feature embeddings or refine inter-image similarity. 
%Shen \emph{et al.}~\cite{shen2018deep} proposed a group-shuffling random walk network for person re-ID by utilizing both probe-gallery and gallery-gallery affinities to improve probe-gallery similarities.
Turcot \emph{et al.}~\cite{turcot2009better}  proposed to augment bag-of-words representations of images by merging of useful features of their neighboring images.
 Iscen \emph{et al.}~\cite{iscen2017efficient} carried out the diffusion through a sparse linear system solver on descriptors of local image regions to refine ranking scores for image retrieval. 
 Donoser \emph{et al.}~\cite{donoser2013diffusion} analyzed a number of diffusion mechanisms and derived a generic framework for iterative diffusion processes in the scope of retrieval applications.
 %Chen \emph{et al.}~\cite{Chen_2018_CVPR} built graph within an image batch, and employed CRF to learn more consistent feature embeddings. 
 Although these methods try to involve more images for feature learning, they need to pre-compute features or can only acquire the features in one image batch. 
 %Luo \emph{et al.}~\cite{luo2018spectral} further refined the obtained features by exploring the top-n gallery images of then ranking list in a post-processing step. However, such post-processing step cannot be optimized in the training procedure, thus is limited by its inflexibility and high cost. 
 To solve this problem, we augment the neural network with an episodic memory module. There also exist some works that added memory modules in the network. 
 Sprechmann \emph{et al.}~\cite{sprechmann2018memorybased} expanded neural network with memory and stored old training examples in memory for parameter adaptation. 
 %Chen \emph{et al.}~\cite{chen2018semi} utilized a memory module to predict the label of unlabelled samples for semi-supervised learning. 
 Sukhbaatar \emph{et al.}~\cite{sukhbaatar2015end} presented a recurrent neural network architecture over a large external memory to allow reading multiple times from memory before outputting a symbol.
 In our case, we retrieve neighbours from the episodic memory module for each training image, and aggregate the neighbourhood information to achieve more discriminative feature embeddings.

\noindent\textbf{Graph Neural Network}. Graph neural network (GNN) is a straightforward extension of CNN from regular Euclidean data to graphs. Following the idea of representation learning, Deep Walk~\cite{perozzi2014deepwalk} is proposed to 
generate graph embedding by combining SkipGram model~\cite{mikolov2013efficient} with graph random walk. 
%Similar approaches like node2vec~\cite{grover2016node2vec} and LINE~\cite{grover2016node2vec} also have achieved excellent performance.
%Instead of learning nodes embedding from a single fixed graph,
GraghSAGE~\cite{hamilton2017inductive} proposed to compute node representations in an inductive manner. It sampled a fixed-size neighbourhood for each node, and then performed simple feature aggregation such as mean pooling, max pooling and LSTM. Different from GraphSAGE that estimates the graph embedding for all the nodes, our method is only interested in the target instance. Then the feature aggregation can be efficiently operated in a tree, and the information is gradually propagated to the root node. To improve the feature aggregation, Petar \emph{et al.}\cite{velivckovic2017graph} proposed an attention-based architecture to perform node classification for graph-structured data, where the attention weights are implicitly learned. Compared with \cite{velivckovic2017graph}, our method fully takes advantages of annotations of images in the memory, and additionally supervises the attention weight with the node-wise class relationships.

\noindent\textbf{Transductive Few-shot Learning}. Given a labelled dataset with different classes, the objective of few-shot learning is to train classifiers for an unseen set of new classes, for which only a few labelled examples are available. Compared with the original few-shot learning problem, transductive few-shot task feeds all test instances simultaneously, which allows us to utilize unlabelled test instances. To solve the problem, TPN~\cite{liu2019fewTPN} spread labels from labelled instances to unlabelled ones with a neighbourhood graph, in which the neighbourhood structure is constructed by a Gaussian similarity function. Instead of performing label propagation, our method aims at enhancing feature embeddings of target nodes by feature aggregation. FEAT~\cite{YeHZS2018Learning} transforms the embeddings from task-agnostic to task-specific by employing self-attention mechanism. In particular, it directly selects related instances by the attention weights, then combined their transformed features to obtain a new feature embedding.  Instead of using linear feature combination, we construct the tree graph with the neighbours retrieved from the memory, then adaptively aggregate the features along the tree to the root node.

\section{Methodology}

\begin{figure*}[t]
	\centering
	\includegraphics[width=1\linewidth]{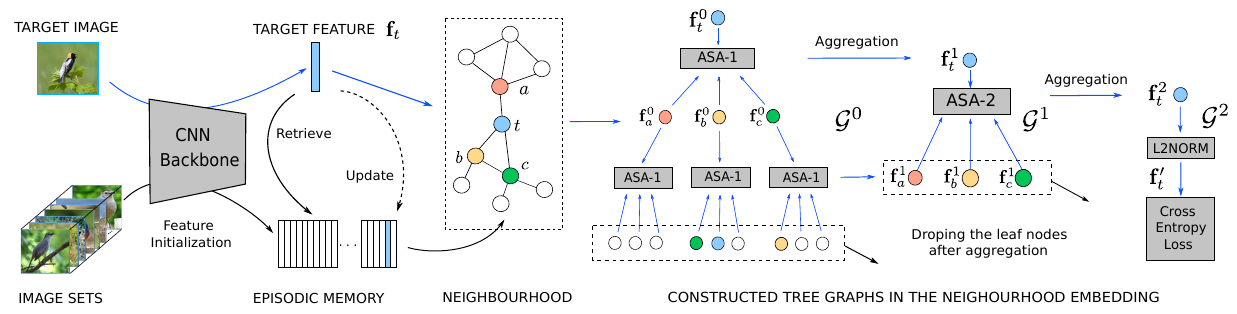}
	\caption{\small The flowchart of the proposed Memory-based Neighbourhood Embedding (MNE). The episodic memory is initialized by the features extracted from a pre-trained CNN backbone. The blue line demonstrates the data flow in training, which includes feature extraction, tree graph construction and aggregative neighbourhood embedding. 
We update the memory with the features from the CNN backbone.}
	\label{fig:pipeline}
\end{figure*}

We aim to enhance the embedding of a single image instance by inspecting the relationships among the instance and its neighbourhood. To efficiently acquire the more relevant neighbours in the feature space,  we augment the neural network with the episodic memory, which provides the necessary features and labels of a large number of instances. Given a target feature extracted from the CNN backbone, we retrieve its neighbours from the memory and organize them to be a tree graph. The final embedding is obtained by an iterative attention-based aggregation strategy over the tree.  In particular, we dynamically prune the leaf nodes of the tree and learn the attention weight by supervising the pairwise relationships in the neighbourhood. The overall framework is illustrated in Fig. \ref{fig:pipeline}.

\begin{algorithm}[t]
 
 \caption{~~~~~~~~~Tree-Graph Construction}
 \renewcommand{\algorithmicrequire}{\textbf{Input:}}
 \renewcommand{\algorithmicensure}{\textbf{Output:}}
 \begin{algorithmic}[1]
 \Require Target node $t$, Memory set $\cM$, tree depth $H$,  neighbour number $K$.
 \Ensure Tree-Graph $\cG(t)$
  \State \parbox[t]{1.0\linewidth}{$\cG(t) = \{t\}$, $h = 0$, $\cL = \{t\}$, where $\cL$ is set of leaf nodes.}
  \vspace{0.3em}
  \While{$h <= H$}
  \For{$v \in \cL $}
  \State $\cN(v) = \Call{SearchNeighbors}{v,K,  \cM}$
  \State $\cG(t) =\Call{AddNewLeafNodes}{v,\cN(v),\cG(t)}$
  \EndFor
  \State $\cL = \Call{GetLeafNodes}{\cG(t)}$
  \State $h = h + 1$
  \EndWhile
  \State \Return $\cG(t)$
 \end{algorithmic}
 \label{alg:tree-graph}
\end{algorithm}

\subsection{Episodic Memory}

Most of the time, the feature embedding of one image can already encode its semantic information for visual recognition. Alternatively, we can also estimate an instance's label by examining its neighbouring features, which is because images with very similar features usually belong to the same or relevant classes. To allow rapid acquisition of more related neighbours while preserving the high performance and good generalization of standard deep models, we add an episodic memory module. 

In training, an episodic memory $\cM$ is composed by both data and label of the training data:
\begin{equation}
 \cM_{train}=\{(\mathbf{f}_{i}, y_{i}) \mid  i \in \mathcal{D}_{train}\},
\end{equation}
where $\mathbf{f}_{i}$ is the feature embedding of an instance $i$, and $y_{i}$ is the associated class label. The number of instances stored in the memory is flexible. Generally, more instances are preferred because they can provide more relevant neighbours. To construct the memory for visual recognition, we first pre-train a CNN feature extractor, then initialize the memory with the extracted features as well as associated labels. Involving the memory supports the end-to-end training, it can be used to search the neighbours over a large number of candidates in the forward pass. As the training influences the CNN backbone, the memory also needs to be updated. We replace the features corresponding to the current training examples in memory by the newly computed features from the CNN backbone.

The label information is not a necessity for model inference in the testing time. To increase the diversity of the memory, we can augment the memory with the unlabelled data in the validation or testing set based on the protocol of specific applications:
\begin{equation}
\cM_{test} = \cM_{train} \cup  \{ \mathbf{f}_{i} \mid  i \in \mathcal{D}_{test/val} \}.
\end{equation}

\begin{algorithm}[t]
 \caption{~~ Aggregative Neighbourhood Embedding}
 \renewcommand{\algorithmicrequire}{\textbf{Input:}}
 \renewcommand{\algorithmicensure}{\textbf{Output:}}
 \begin{algorithmic}[1]
 \Require  Tree-Graph $\cG(t)$, tree depth $H$.
 \Ensure Neighbourhood embedding $\mathbf{f}'_t$ of $\cG(t)$.
\State $\cG^0 = \cG(t), h = 0$
\While{$h < H$} \\
// branch node means a node with at least one child
\State $\cB^{h} = \Call{GetBranchNodes}{\cG^{h}} $
\For{$v \in \cB^{h} $}
\State $\cC(v) = \Call{GetChildNodes}{v,\cG^{h}}$
\State $\mathbf{f}^{h+1}_{v} = \Call{Aggregate}{\mathbf{f}^{h}_{v}, \cC(v)}$
\EndFor
\State $\cL = \Call{GetLeafNodes}{\cG^{h}}$
\State $\cG^{h+1} = \cG^{h} \setminus \cL$     \;\;\;\;  // drop leaf nodes
\State $h = h +1$
\EndWhile
  \State $\mathbf{f}'_t = \Call{L2Norm}{\mathbf{f}^H_{t}}$  \;\;\;\;  // normalization
  \State \Return $\mathbf{f}'_t$
 \end{algorithmic}
 \label{alg:embedding}
\end{algorithm}

 \subsection{Embedding with Neighbourhood Tree Graph}

The main purpose of our method is to obtain a more robust feature embedding for visual recognition.  Instead of extracting the feature from a single image, we ``re-estimate" the feature from its neighbourhood, which is modelled by a tree graph. In particular, we take the target instance as the root node, then build the tree in an iterative fashion.  Each time, we extend all the leaf nodes by adding their $K$ nearest neighbours from the memory $\mathcal{M}$ as the new leaf nodes. The tree graph grows until it achieves a predefined depth $H$. The detailed procedure is demonstrated in Alg. \ref{alg:tree-graph}.  It is noteworthy that we allow a same instance appear multiple times in the tree. The frequently appeared nodes are usually ``center" instances in the neighbourhood, which will have a high influence on the final feature embedding.

The embedding of an instance then can take advantages of neighbourhood tree graph to exploit more abundant information. As the nodes in the graph do not have ordering information, thus the standard neural networks like CNNs or RNNs cannot be directly adopted. To handle the graph input, we iteratively perform feature aggregation among connected nodes, which gradually propagates information within the graph to the target instance. Specifically, the $h$-th feature aggregation for node $u$ can be represented by:
\begin{equation}
\mathbf{f}^{h}_{u} = \Call{Aggregate}{\mathbf{f}^{h-1}_{u}, \cC(u)}, \label{Eq:Agg}
\end{equation}
where $\mathcal{C}(u)$ is the child nodes of node $u$ and $\mathbf{f}^{0}_{u}$ is initialized by the original feature. Intuitively, the nodes close to the target instance should have more influence over the final embedding. For this reason, we perform the feature aggregation over a dynamic graph as demonstrated in Alg. \ref{alg:embedding}. Each time, we update the features of all the branch nodes in the tree, then drop the leaf nodes to form a new tree. Finally, the tree remains only one node whose feature is the neighbourhood embedding. We impose the cross entropy loss over the neighbourhood embeddings: 
\begin{equation}
\mathcal{L}_{C}(t)=-\sum_{i=1}^{I} y_{i,t} \log \left( \frac{\exp(\mathbf{w}_{i}^{\top} \mathbf{f}'_{t} ) }{\sum_{j=1}^{I} \exp(\mathbf{w}_{j}^{\top}\mathbf{f}'_{t})} \right), \label{Eq:CE}
\end{equation}
where $y_{i,t}$ is the index label with $y_{i,t}=1$ if the image $t$ belongs to
the $i$th class and $y_{i,t}=0$ otherwise. There are $I$ classes in total.

\subsection{Aggregation with Supervised Attention}

General aggregation strategies like mean-pooling and max-pooling cannot determine which neighbours are more important. To adaptively aggregate the features from the same class, which is crucial for visual recognition, we propose a network module named ASA to \textbf{A}ggregate features with \textbf{S}upervised \textbf{A}ttention.

\begin{figure}[t]
\setlength{\abovecaptionskip}{2pt}
	\centering
	\includegraphics[width=1\linewidth]{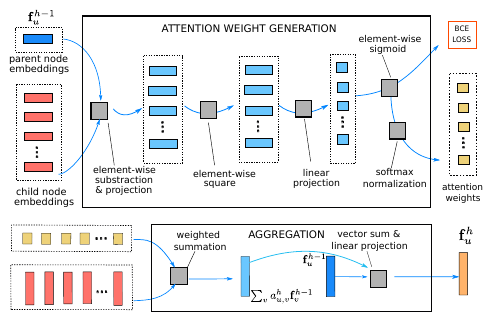}
	\caption{\small Illustration of the proposed ASA-$h$ module. We estimate the current feature of a node by aggregating the previous feature and the features of its child nodes. The attention weights are supervised by whether two nodes belong to the same class.} \label{fig:attention}
\end{figure}

The $h$-th aggregation of all the parents nodes in the graph is accomplished by a same module (as in Fig.~\ref{fig:pipeline}), denoted by ASA-$h$. In the module, we introduce attention weights over the child nodes, then aggregation is specified by: 
\begin{equation}
\mathbf{f}^{h}_u = \mathbf{W}^{h}_{A}\big(\mathbf{f}^{h-1}_{u} + \textstyle{\sum_{v \in \mathcal{C}(u)}} a^{h}_{u,v} \mathbf{f}_{v}^{h-1} \big) + \mathbf{b}^{h}_{A},
\end{equation}
where $\mathbf{W}^{h}_{A}, \mathbf{b}^{h}_{A}$ are parameters for feature transformation, and $a^{h}_{u,\ast}$ are attention weights. The feature embedding of child node $\mathbf{f}_{v}^{h-1}$ needs to be mapped close to that of the  parent node $\mathbf{f}_{u}^{h-1}$ if they are from the same class, therefore, the attention weight $a^{h}_{u,v}$ needs to be high. Different from most approaches that implicitly learn the attention weights in the network, we utilize the label information in the memory to supervise the attention module. In particular, the attention weight is designed to be proportional to the probability of two nodes belonging to the same class:
\begin{equation}
   a^{h}_{u,v} =  p^{h}_{u,v} / \sum_{k \in \mathcal{C}(u)} p^{h}_{u,k}.
\end{equation}
The probability $p^{h}_{u,v}$ is estimated from the previous feature embeddings $\mathbf{f}_{u}^{h-1}$ and $\mathbf{f}_{v}^{h-1}$ by the following steps:
\begin{equation}\small
\begin{split}
	 \mathbf{d}^{h}_{u,v}  & = \mathbf{W}^{h}_{D}(\mathbf{f}_{u}^{h-1} \!-\! \mathbf{f}_{v}^{h-1})+\mathbf{b}^{h}_{D},  \\
	 p^{h}_{u,v}  & = \sigma(\mathbf{W}^{h}_{S} (\mathbf{d}^{h}_{u,v} \circ  \mathbf{d}^{h}_{u,v}) + \mathbf{b}_{S}^{h}),
\end{split}
\end{equation}
where $\mathbf{W}_{D}^{h}$, $\mathbf{b}_{D}^{h}$ and $\mathbf{W}_{S}^{h}, \mathbf{b}_{S}^{h}$ are parameters of linear transformations to obtain the difference feature vector and pairwise probability. With $\mathbf{d}^{h}_{u,v}$, we first perform element-wise square, then project the obtained vector to a scalar value, and finally normalize the scalar value to be within (0,1) with the sigmoid function $\sigma$. We supervise the probabilities between the parent nodes and all its child nodes:
\begin{equation}
\small{  \mathcal{L}_{P}^{h}(u) = -\textstyle{\sum_{v \in \mathcal{C}(u)} [y_{u,v} \log p^{h}_{u,v} + (1-y_{u,v})\log (1- p^{h}_{u,v}) ]},}  \label{Eq:BCE}
\end{equation}
where $u \in \mathcal{B}^{h}$ which $\cB^{h}$ is set of branch nodes of $h$th tree graph. $y_{u,v}=1$ if nodes $u$ and $v$ belong to the same class, otherwise $y_{u,v}=0$. Fig. \ref{fig:attention} shows the detailed architecture of the proposed ASA-$h$ module.

In total, our MNE is learned with two kinds of loss functions. One is the multi-class cross-entropy loss (Eq.~\ref{Eq:CE}), which is imposed over the neighbourhood embeddings, the other is the binary cross-entropy loss (Eq.~\ref{Eq:BCE}), which supervises the pairwise probabilities between nodes in each tree. 

\section{Implementation}

 The proposed Memory-based Neighbourhood Embedding (MNE) is applied on two recognition tasks, $\ie$, image search and transductive few-shot learning. We now introduce the CNN backbone selection, the memory construction and the training details for the two tasks accordingly.

\noindent \textbf{Image Search.}  Following the practice in~\cite{sun2018beyond}, we adopt ResNet-50~\cite{he2016deep} as the backbone network, and change the stride of the last down sampling block from 2 to 1, which is helpful to get more fine-grained features. We pre-train the CNN backbone to classify all the classes in the training set, then use the pre-trained network to extract the features of the entire training set and use them to initialize the episodic memory in the training phase. 

In training, we augment the input images by random horizontal flipping and random erasing~\cite{zhong2017random}, then employ Adam~\cite{kingma2014adam} to optimize the entire model. As the CNN backbone has been pre-trained, the initial learning rates are $10^{-5}$ for CNN backbone and $10^{-4}$ for other parts of the model, respectively.  We decay the learning rates by 0.1 after 20 epochs, and obtain the final model after 40 epochs. In testing, we utilize the CNN backbone to extract the features of gallery images, and augment them into the memory. With this memory, we estimate the neighbourhood embeddings for all the query and gallery images, and utilize the embeddings to perform the image search.

\noindent \textbf{Transductive Few-shot Learning.}  For a fair comparison with the existing methods, we employ a widely-used four-layer convolution network \cite{finn2017model, snell2017prototypical} as the backbone. It contains 4  blocks. Each block has a convolutional layer with kernel size 3, a batch normalization layer, a relu activation layer, and a max pooling layer. We pre-train the backbone network with the cross-entropy loss over the training set.

The training follows the episodic strategy. We mimic the N-way M-shot tasks in each training batch in order to handle the N-way M-shot tasks in testing. The memory contains the feature embeddings for all the training and testing images in the current episode.  \emph{E.g.}, consider a typical 1-shot 5-way task with 15 testing instances per-class, the memory will have 80 (5+5x15) images. With a pre-trained CNN-backbone~\cite{YeHZS2018Learning}, we still employ Adam for optimization, where the initial learning rates for CNN backbone and other parts are $10^{-4}$ and $10^{-3}$. The learning rate is decayed by 0.1 every 5000 episodes and the training is stopped after 30000 episodes. In testing, we first estimate the neighbourhood embeddings for each image in the memory, then assign labels to the testing images according to their affinities to the training images.

\section{Experiments}

We evaluate the proposed method on four datasets. Ablation studies are mainly conducted on CUHK03~\cite{li2014deepreid} and \emph{mini}ImageNet~\cite{vinyals2016matching}, which are about person search and few-shot learning, respectively. Besides, we report more results on DukeMTMC \cite{ristani2016performance} and \emph{tiered}ImageNet~ \cite{ren18fewshotssl} to compare with the current state-of-the-art methods.

\begin{figure}[t]
	\centering
	\includegraphics[width=1\linewidth]{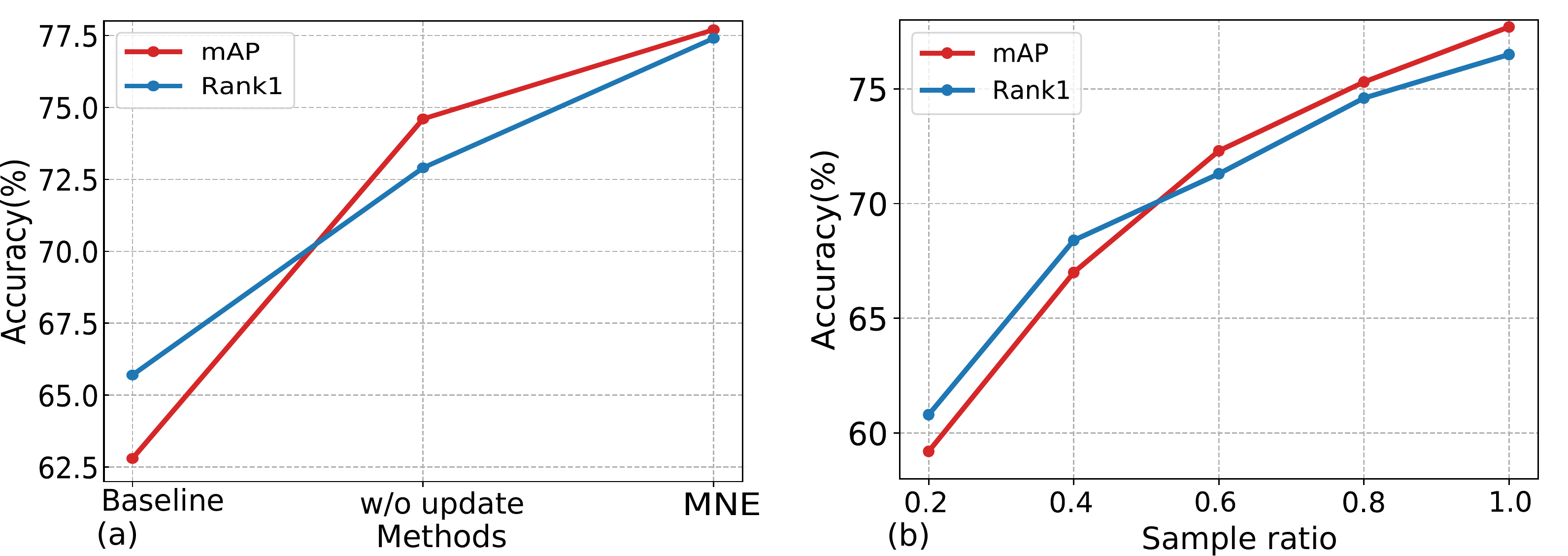}
	\caption{\small Investigation of the episodic memory. We evaluate (a) the effectiveness of memory update strategy; (b) the influence of memory size in terms of mAP and rank-1 accuracies.} \label{fig:memory}
	\vspace{1.2em}
	\centering
	\includegraphics[width=1\linewidth]{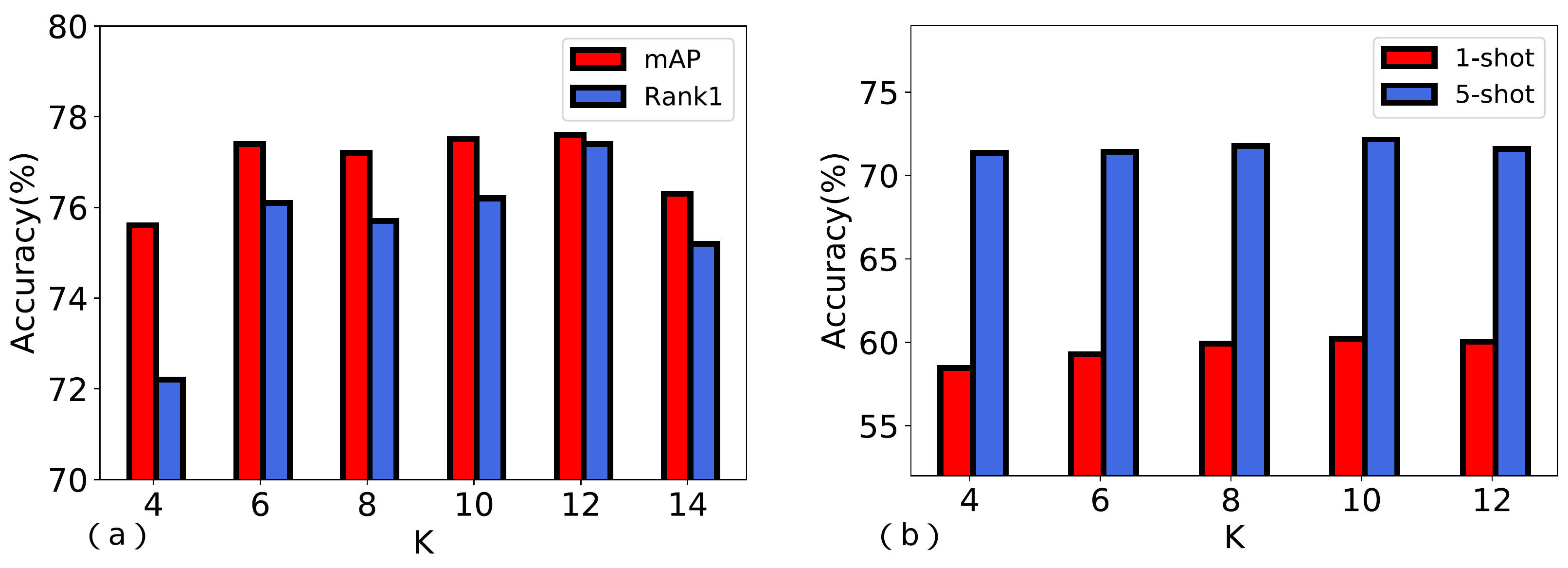}
	\caption{\small Influnence of nearest neighbour number $K$. We report the mAP and rank-1 accuracy on CUHK03, and report the 1-shot and 5-shot classification accuracies on \emph{mini}Imagenet.} \label{fig:tree_param}
\end{figure}

\subsection{Experimental Setup}

\noindent \textbf{Datasets}. CUHK03 and DukeMTMC are two large-scale person search benchmarks. CUHK03 contains 14,096 images of 1,467 identities. Each identity is captured from two cameras and has an average of 4.8 images in each camera. We follow the training/testing protocol proposed in~\cite{zhong2017re}, which splits the dataset into a training set with 767 identities and a testing set with 700 identities. DukeMTMC is a subset of the multi-target, multi-camera pedestrian tracking dataset \cite{ristani2016performance}. It contains 1,812 identities captured by 8 cameras. There are 36,411 images in total, where 16,522 images of 702 identities are used for training, 2,228 images of another 702 identities are used as query images, and the remaining 17,661 images are gallery images. For few-shot learning, \emph{mini}ImageNet~\cite{vinyals2016matching} and \emph{tiered}ImageNet~\cite{ren18fewshotssl} are two widely-used benchmarks. \emph{mini}ImageNet has 60, 000 images of 100 classes selected from the ILSVRC-12 dataset~\cite{russakovsky2015imagenet}, and each class has 600 images. Following the protocol in~\cite{ravi2016optimization}, we use 64 classes for training, 16 classes for validation, and 20 classes for testing. \emph{tiered}ImageNet ~\cite{ren18fewshotssl} is also a subset of ImageNet, but it has 608 classes much larger than that in \emph{mini}ImageNet. All the classes are summarized into 34 categories, which are further divided into 20 training (351 classes), 6 validation (97 classes) and 8 test (160 classes) categories. Such strategy ensures that the training classes are distinct from the test classes. It is a more challenging and realistic few-shot setting.

\noindent \textbf{Evaluation Metric}.  The cumulated matching accuracy at rank-1 and the mean average precision (mAP) are adopted for image search. We 
evaluate 1-shot 5-way and 5-shot 5-way classification tasks with 10,000 sampled test episodes for few-shot learning, and report the mean accuracy and the 95\% confidence interval.

\begin{figure}[t]  
\begin{minipage}{0.46\textwidth}
  \centering 
  \setlength{\tabcolsep}{2.5mm}{
    \begin{tabular}{lcccc}  
    \toprule  
    \multirow{2}{*}{Search depth}&  
    \multicolumn{2}{c}{CUHK03}&\multicolumn{2}{c}{$\emph{mini}$ImageNet}\cr  
   \cmidrule(lr){2-3} \cmidrule(lr){4-5}
    &mAP&rank1&1-shot&5-shot\cr  
    %\midrule  
    \hline \hline
    0 (baseline) & 62.8 & 65.7 & 50.41 & 70.52      \cr
    1 & 75.7 & 75.5 & 59.68 & 71.71\cr  
   % \hline
    2 &\textbf{77.7} & \textbf{77.4} &  \textbf{60.20} & \textbf{72.16} \cr  
     % \hline
    3 &76.9 & 76.2 &  59.83 & 71.12 \cr 
    \bottomrule  
    \end{tabular}  }
    \captionof{table}{\small The influence of search depth $H$. We fix $K\!=\!12$ for CUHK03 and $K\!=\!10$ for $\emph{mini}$ImageNet. The 95\% confidence interval was omitted for simplicity.} \label{tab:H_vary}
    \vspace{1em}
\end{minipage}

\begin{minipage}{0.48\textwidth}
\includegraphics[width=1.0\linewidth]{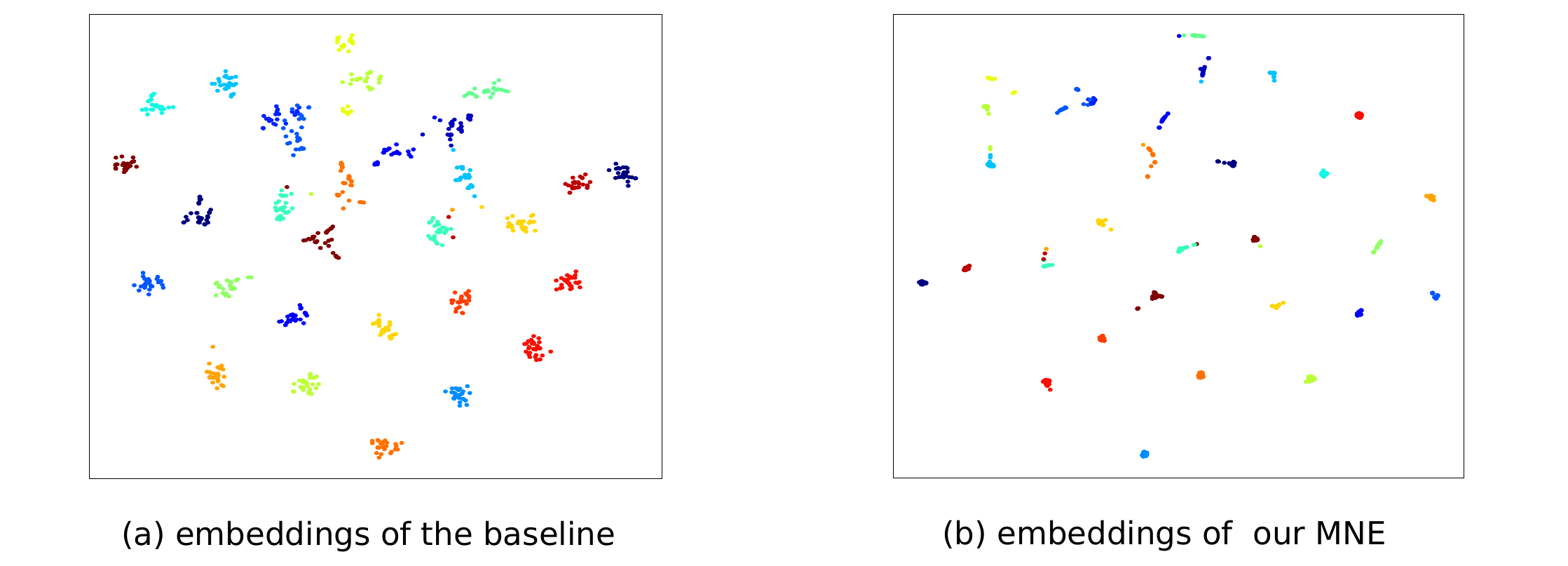}
\caption{\small  t-SNE visualization of feature embeddings. Each point indicates a testing image from randomly selected 30 identities of DukeMTMC. Different identities may share the same color.}
\label{fig:vis_image}
\end{minipage}
\end{figure}

\subsection{Ablation Studies}
 \begin{figure*}[t]
 \setlength{\belowcaptionskip}{-0.1pt}
 \setlength{\abovecaptionskip}{0pt}
  \centering
  \subfigure[{ mAP on CUHK03}]{
    \label{fig:subfig:attn_ave_map} %% label for first subfigure
    \includegraphics[width=0.23\linewidth]{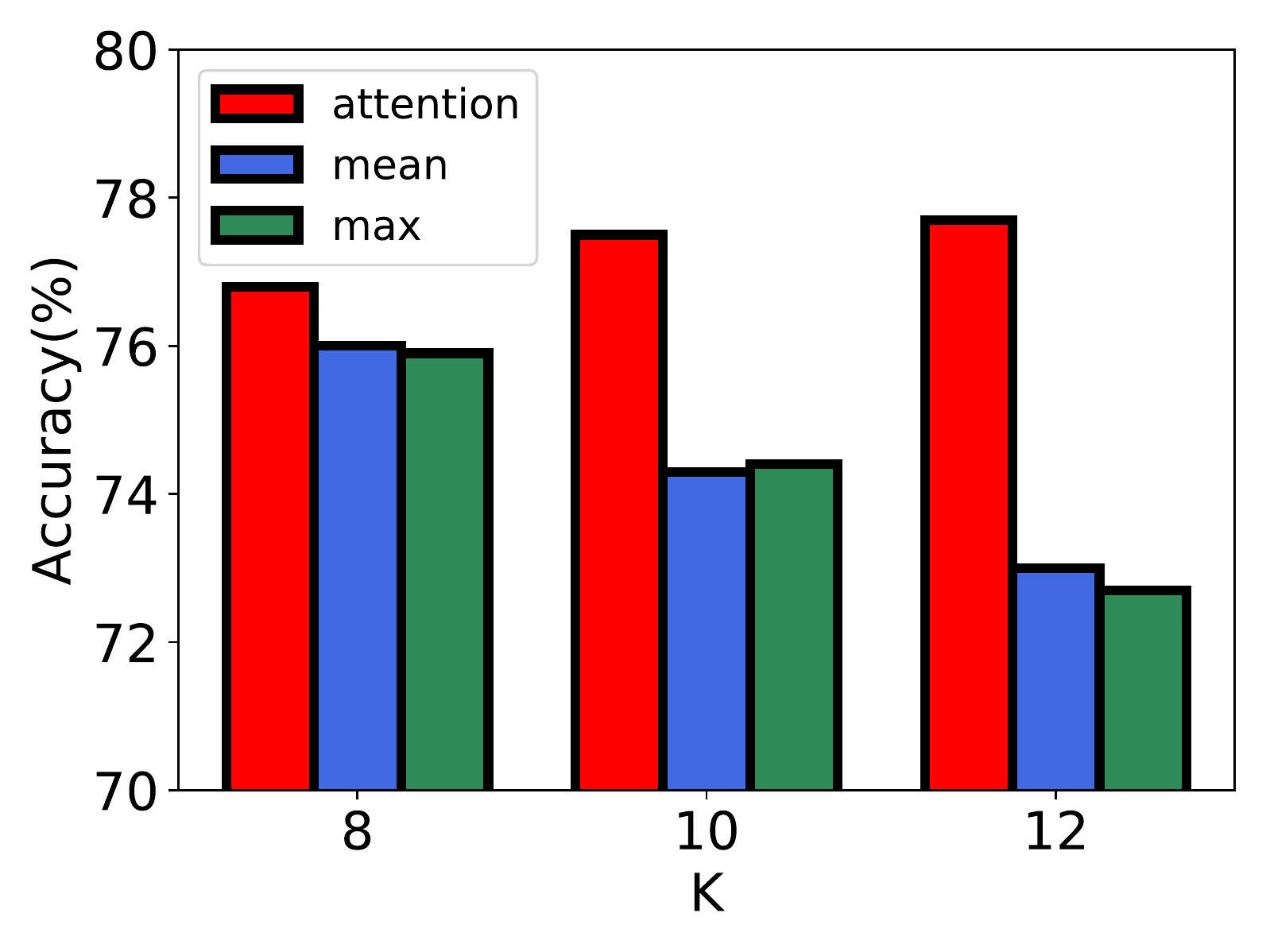}}
  %\hspace{1in}
  \subfigure[Rank-1 Acc. on CUHK03]{
    \label{fig:subfig:attn_ave_rank1} %% label for second subfigure
    \includegraphics[width=0.23\linewidth]{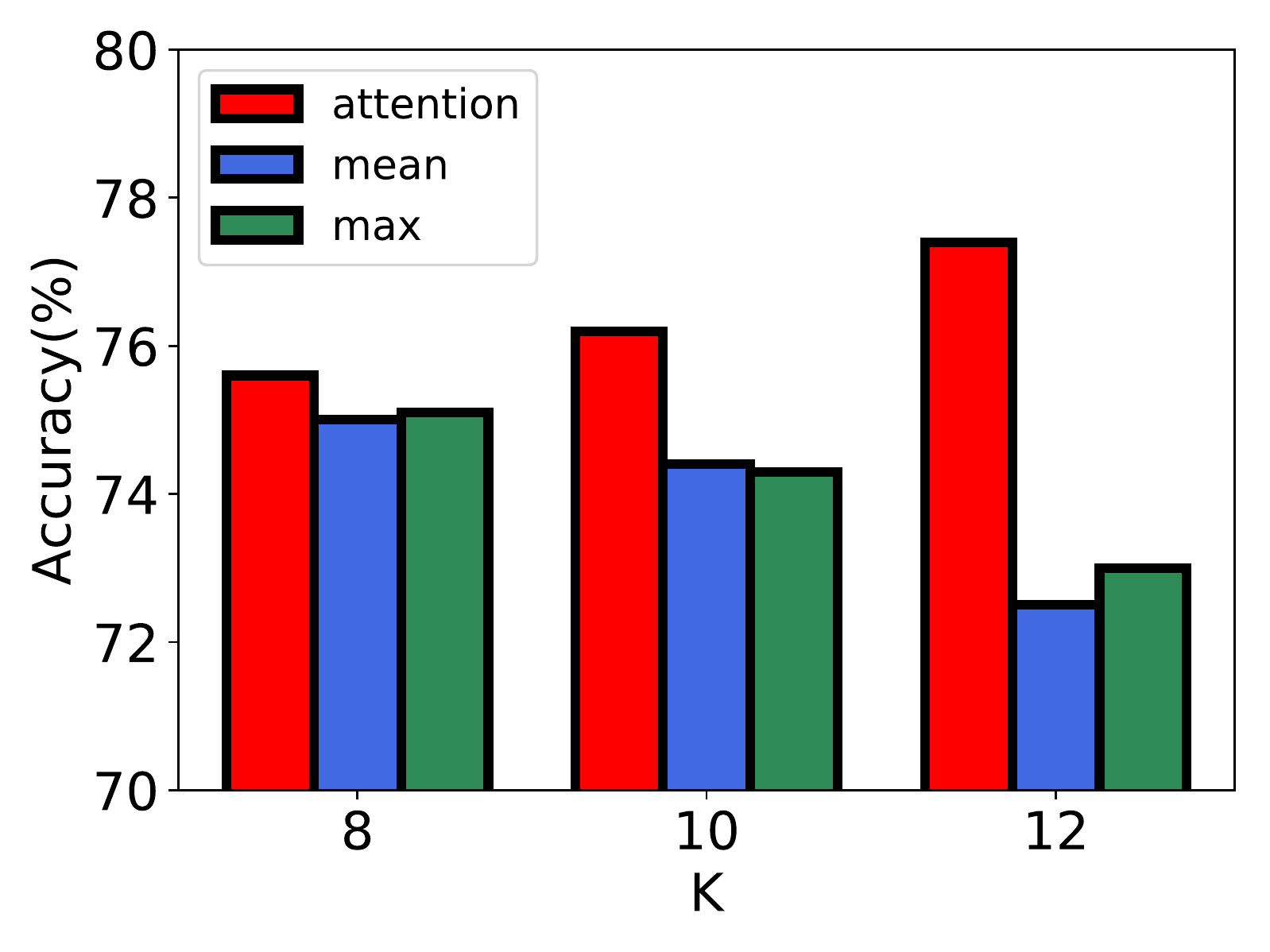}}
    \subfigure[{ 1-shot Acc. on $\emph{mini}$ImageNet}]{
    \label{fig:subfig:attn_ave_1_shot} %% label for first subfigure
    \includegraphics[width=0.23\linewidth]{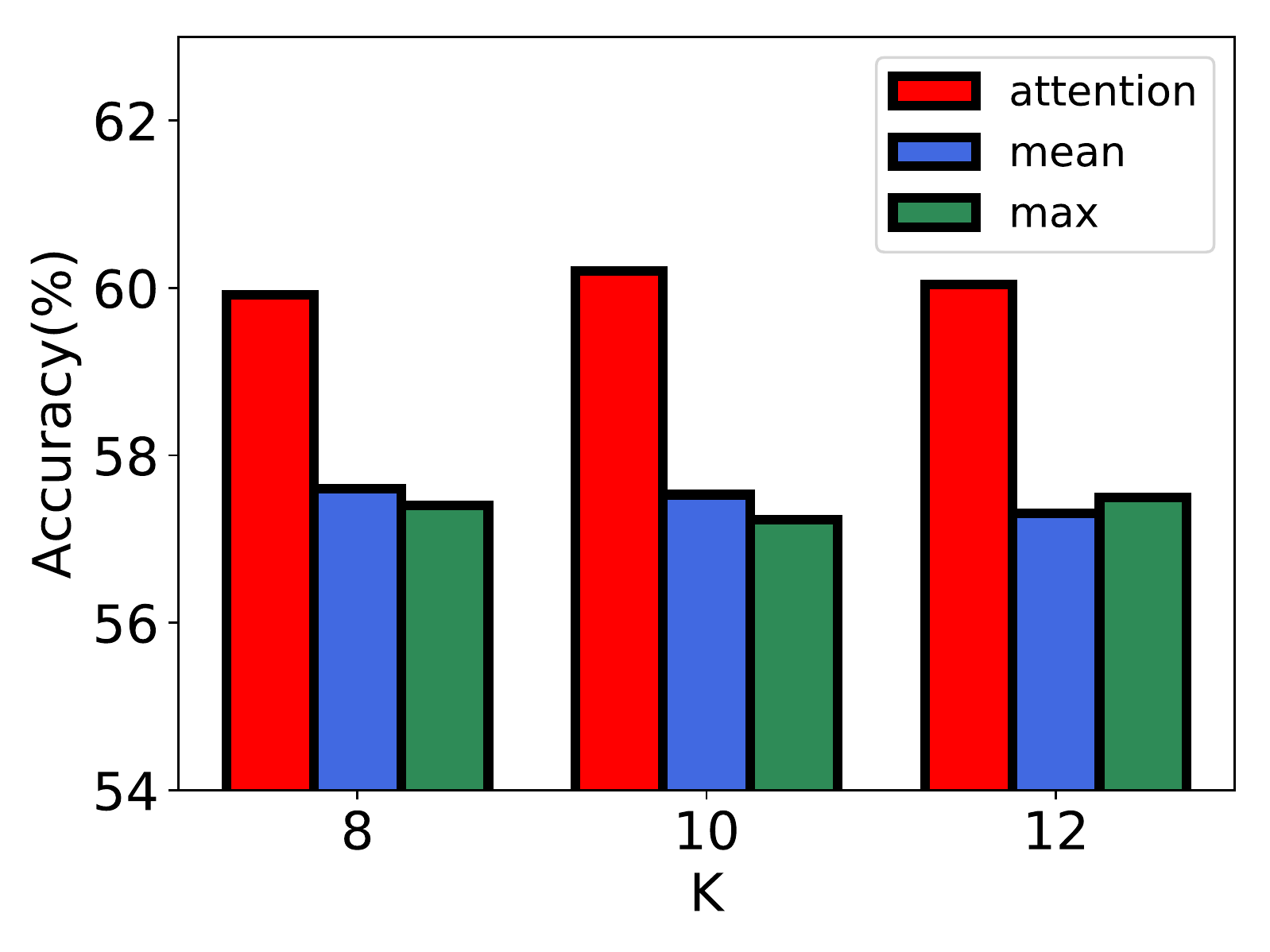}}
  %\hspace{1in}
  \subfigure[5-shot Acc. on $\emph{mini}$ImageNet]{
    \label{fig:subfig:attn_ave_5_shot} %% label for second subfigure
    \includegraphics[width=0.23\linewidth]{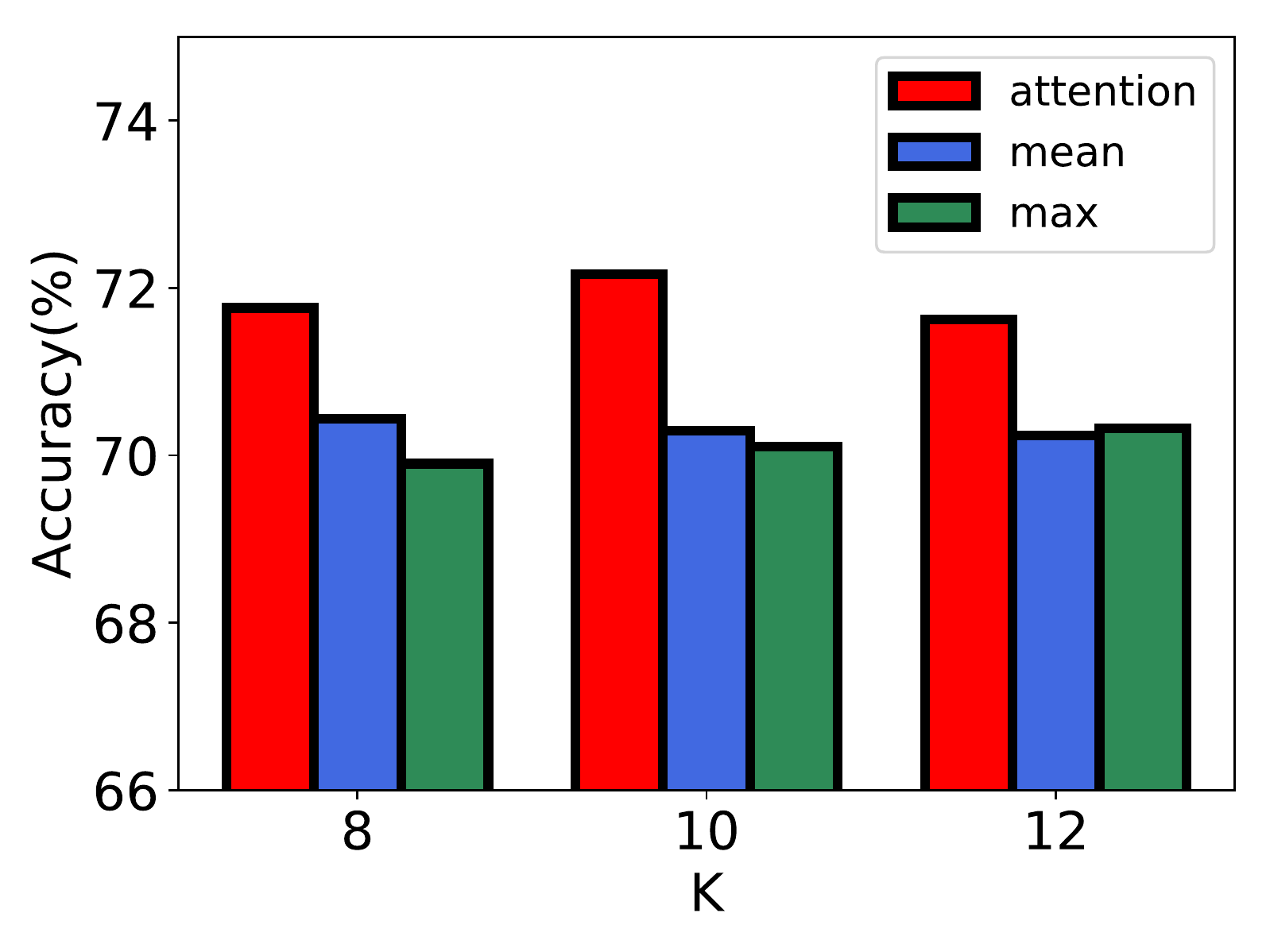}}
  \caption{\small The effectiveness of attention aggregation on CUHK03 for image search and \emph{mini}ImageNet for transductive few-shot learning.}
  \label{fig:attn_ave} %% label for entire figure
\end{figure*}

We investigate the main components of the proposed MNE, including the episodic memory, tree graph embedding and supervised attention. 

\subsubsection{Memory}

Incorporating the episodic memory for feature embeddings is a characteristic of our method and the prerequisite to perform the neighbourhood embedding.  We investigate the influence of the memory size and the memory update strategy on the CUHK03 dataset, which can provide memory with flexible size in both training and testing.

\noindent  \textbf{Memory Update.} As the CNN backbone changes in the training, the memory update is an important step to obtain the updated features for neighbourhood embedding. We compare three variants, the baseline, the MNE without memory updating and the proposed MNE. Among them, the baseline method indicates directly extracting  features  from the CNN backbone. The results in Fig. \ref{fig:memory}a shows that the MNE without memory update has already significantly improved the baseline, and updating the memory can bring additional gain. It improves the rank-1 accuracy from 72.9\% to 77.3\% and the mAP from 74.6\% to 77.7\%.

\noindent \textbf{Memory Size.} In testing, we sample different number of gallery images to construct the memory. The Fig. \ref{fig:memory}b reports the performance varying with the sampling ratios. It can be seen that larger sampling ratio,  \emph{i.e.},  more instances in the memory leads to better performance, which verifies the assumption that larger memory can provide more relevant neighbours to improve the neighbourhood embeddings.

\subsubsection{Tree Graph Neighbourhood Embedding}

We propose tree-graph based embedding network to exploit the context information in the feature space, aiming to enhance the original CNN features. As the tree graph structure is important to the embedding, we first study how tree structure can influence the performance, then compare the obtained neighbourhood embeddings with the CNN features.
 
\noindent \textbf{Tree Construction.}  We study the number of neighbours $K$ and the tree depth $H$ in tree construction. To build the tree graph, we extend the leaf nodes by adding their $K$ nearest neighbours from the memory.  We observe how the performance varies with $K$ on CUHK03 and \emph{mini}Imagenet when $H=2$. The results in Fig. \ref{fig:tree_param} show that too large and too small $K$ will lead inferior results. This is because too small $K$ will not get sufficient neighbours while too large $K$ will  introduce unrelated neighbours which may weaken the effectiveness of feature aggregation. We also observe how $H$ influence the performance by fixing $K=12$ in Tab. \ref{tab:H_vary}. Best results are achieved when $H=2$, and deeper tree graph will not bring additional gain. As the search depth increase, more unrelated samples may be introduced, which possibly impairs the feature embeddings of the target samples.

\begin{table}[t]  
  \centering \small
 \setlength{\tabcolsep}{2.7mm}
 {
    \begin{tabular}{c|c|c|c|c}  
    %\toprule  
    \hline
	Dataset & K & metric & with BCE & w/o BCE\cr
    \hline\hline 
    \multirow{2}{*}{CUHK03} & \multirow{2}{*}{8} & mAP & 76.8 & 76.0\cr  
    & & rank1 &  75.6 & 75.0 \cr  
    \hline
    \multirow{2}{*}{CUHK03} & \multirow{2}{*}{12} & mAP &  77.7 &73.0\cr   
    & & rank1 &  77.4 & 72.5 \cr  
     \hline
    \multirow{2}{*}{ \emph{mini}ImageNet} & \multirow{2}{*}{8} & 1-shot &  59.92 &58.51\cr   
    & & 5-shot &  71.76 & 71.40 \cr  
     \hline
    \multirow{2}{*}{ \emph{mini}ImageNet} & \multirow{2}{*}{12} & 1-shot &  60.04 &59.32\cr   
    & & 5-shot &  71.62 & 71.54 \cr  
    %\bottomrule  
    \hline
    \end{tabular}  }
    \vspace{0.5em}
    \caption{
    \small Effectiveness of BCE loss of attention on CUHK03 and  \emph{mini}ImageNet.} 
     \label{tab:attention_BCE}  
    \vspace{-1em}
\end{table}

\noindent \textbf{Neighbourhood Embedding v.s. Backbone Feature.}  The performance gap between the two kinds of features embeddings can be reflected in Tab.~\ref{tab:H_vary},  where our method can outperform the baseline by a large margin on both image search and few-shot learning tasks.  The 1-shot scenario can benefit more from the neighbourhood embedding than the 5-shot scenario, which is consistent with the results in~\cite{liu2019fewTPN}.  With the shot increase, more labelled images are available in the testing phase, thus the effectiveness of neighbourhood embedding, \emph{i.e.}, exploiting the unlabelled context, will be weaken. In addition, we employ t-SNE to visualize the feature embeddings of the same 30 testing persons from DukeMTMC by CNN backbone and neighbourhood embedding in Fig. \ref{fig:vis_image}, which clearly shows that incorporating neighbourhood embedding can generate more discriminative feature embeddings. 

\subsubsection{Supervised Attention}

\noindent \textbf{Effectiveness of the Attentive Aggregation.}  We compare the proposed attentive aggregation with the mean/max feature pooling methods, which are most straightforward strategies to summarize features. Results on CUHK03 and \emph{mini}ImageNet are displayed in Fig.~\ref{fig:attn_ave}, where the attentive aggregation outperforms the mean/max aggregation on both tasks. With the increase of $K$, the results of mean/max aggregation become worse while our method is stable.

\noindent \textbf{Effectiveness of the Attention Supervision.} We impose the BCE loss (Eq. \ref{Eq:BCE}) within the ASA module, in order to make the attention weight can reflect whether two images belong to a same class. The results in Tab.~\ref{tab:attention_BCE} show that imposing BCE can generally achieve superior performance.

%-----------------------------------------------------------------------------

\begin{table}[t]  
  \centering \small
 \setlength{\tabcolsep}{4mm}
 {  
   \begin{tabular}{l|c|c|c}  
   % \toprule   
   \hline
    Method &Ref&mAP&rank1\cr  
    %\midrule 
    \hline \hline
   HA-CNN~\cite{li2018harmonious}&CVPR'18  & 63.8 & 80.5\cr
    MLFN~\cite{chang2018multi}&CVPR'18&62.8 & 81.0  \cr  
    DuATM~\cite{si2018dual}&CVPR'18&64.6 & 81.8 \cr   
   PCB~\cite{sun2018beyond}&ECCV'18&69.2 & 83.3 \cr   
    Part-aligned~\cite{suh2018part}&ECCV'18&69.3&84.4 \cr  
    Mancs~\cite{wang2018mancs} &ECCV'18&71.8&84.9 \cr 
    GSRW~\cite{shen2018deep}&CVPR'18  & 66.4 & 80.7\cr
    SGGNN~\cite{shen2018person}&ECCV'18&68.2&81.1 \cr 
    Spectral~\cite{luo2018spectral}&Arxiv'18 &73.2&86.9 \cr
    Spectral+post~\cite{luo2018spectral}&Arxiv'18 &79.6&90.0 \cr
    Proposed MNE &&\textbf{87.5}&\textbf{90.4}\cr
    %\bottomrule  
    \hline
    \end{tabular}  
   }
    \caption{\small Experimental results of the proposed MNE and state-of-the-art methods on the DukeMTMC dataset. } 
     \label{tab:sota_on_DukeMTMC}  
    \vspace{1em}
  \centering \small
 \setlength{\tabcolsep}{4mm}
 {  
   \begin{tabular}{l|c|c|c}  
    %\toprule   
    \hline
    Method &Ref&mAP&rank1\cr  
    %\midrule 
    \hline \hline
    SVDNet~\cite{sun2017svdnet}&ICCV'17  & 37.8 & 40.9\cr
    DPFL~\cite{chen2017person}&ICCV'17&40.5 & 43.0  \cr  
    HA-CNN~\cite{li2018harmonious}&CVPR'18&41.0 & 44.4 \cr   
    MLFN~\cite{chang2018multi}&CVPR'18&49.2 & 54.7 \cr   
    DaRe~\cite{wang2018resource}&CVPR'18&61.6 & 66.1 \cr   
    SFT~\cite{luo2018spectral}&Arxiv'18 &62.4&68.2 \cr
    SFT+post~\cite{luo2018spectral} &Arxiv'18 &71.7&74.3 \cr
    Proposed MNE &&\textbf{77.7}&\textbf{77.4}\cr
    %\bottomrule  
    \hline
    \end{tabular}  
    }
    \caption{\small Experimental results of the proposed MNE and state-of-the-art methods on the CUHK03 dataset.} 
     \label{tab:sota_on_cuhk03}  
    \vspace{-1em}
\end{table}

\begin{table}[t]  
  \centering \small
 \setlength{\tabcolsep}{1.5mm}
 {  
    \begin{tabular}{l|c|c|c}  
    %\toprule  
    \hline  
    Method &Ref&1-shot&5-shot\cr  
    %\midrule 
    \hline \hline
    MAML~\cite{vinyals2016matching}&NIPS'16  & $48.70_{\pm1.84}$ & $63.11_{\pm0.92}$\cr
    ProteNet~\cite{snell2017prototypical}&NIPS'17& $46.14_{\pm0.77}$ &$65.77_{\pm0.70}$  \cr  
    RelationNet~\cite{sung2018learning}&CVPR'18& $51.38_{\pm0.82}$ & $67.07_{\pm0.69}$ \cr   
    Feat~\cite{YeHZS2018Learning}&Arxiv'18&$55.21_{\pm0.20}$ &$72.17_{\pm0.16}$ \cr 
    \hline 
    Semi-ProtoNet~\cite{ren18fewshotssl}&ICLR'18& $50.41_{\pm0.31}$ & $64.59_{\pm0.28}$\cr
    TPN~\cite{liu2019fewTPN}&ICLR'19 &$53.75_{\pm0.86}$ & $69.43_{\pm0.67}$ \cr
    TPN+higher shot~\cite{liu2019fewTPN} &ICLR'19 & $55.51_{\pm0.86}$ & $69.86_{\pm0.65}$ \cr
    FEAT+transductive~\cite{YeHZS2018Learning}&Arxiv'18 &$56.49_{\pm0.21}$ & $\textbf{72.65}_{\pm0.16}$ \cr
    Proposed MNE &&$\textbf{60.20}_{\pm0.23}$& $72.16_{\pm0.17}$\cr
    %\bottomrule  
    \hline
    \end{tabular}  
    }
    \caption{\small Experimental results of the proposed MNE and state-of-the-art methods on $\emph{mini}$ImageNet. The results of original few-shot learning and transductive few-shot learning methods are separated.} 
   \label{tab:sota_on_mini}  
 \vspace{2em}
  \centering \small
 \setlength{\tabcolsep}{1.5mm}
 {  
    \begin{tabular}{l|c|c|c}  
    %\toprule   
    \hline
    Method &Ref&1-shot&5-shot\cr  
   % \midrule 
    \hline \hline
    MAML~\cite{vinyals2016matching}&NIPS'16  & $51.67_{\pm1.81}$ & $70.30_{\pm1.75}$\cr
    ProteNet~\cite{snell2017prototypical}&NIPS'17& $48.58_{\pm0.87}$ &$69.57_{\pm0.75}$  \cr  
    RelationNet~\cite{sung2018learning}&CVPR'18& $54.48_{\pm0.93}$ & $71.31_{\pm0.78}$ \cr   
    \hline 
    Semi-ProtoNet~\cite{ren18fewshotssl}&ICLR'18& $52.39_{\pm0.44}$ & $70.25_{\pm0.31}$\cr
    TPN~\cite{liu2019fewTPN}&ICLR'19 &$57.53_{\pm0.96}$ & $72.85_{\pm0.74}$ \cr
    TPN+higher shot~\cite{liu2019fewTPN} &ICLR'19 & $59.91_{\pm0.94}$ & $73.30_{\pm0.75}$ \cr
    Proposed MNE&&$\textbf{60.04}_{\pm0.28}$& $\textbf{73.63}_{\pm0.21}$\cr
    %\bottomrule 
    \hline 
    \end{tabular}  }
    \caption{\small  Experimental results of the proposed MNE and state-of-the-art methods on $\emph{tired}$ImageNet. The results of original few-shot learning and transductive few-shot learning methods are separated.} \label{tab:sota_on_tired}  
   \vspace{-1em}
\end{table}

\subsection{Comparison with State-of-the-art Approaches}

\noindent \textbf{Image Search.}   We report the comparison results between our method and the state-of-the-art approaches in Tab.~\ref{tab:sota_on_DukeMTMC} and Tab.~\ref{tab:sota_on_cuhk03} for DukeMTMC and CUHK03, and our method significantly outperforms the others without any additional post-processing. Our method achieves 87.5\% and 77.7\% mAP on DukeMTMC and CUHK03, which improve the competitive Spectral+post\cite{luo2018spectral} by  7.9\% and 6.0\%, respectively. Spectral+post and our method both exploit context information in testing, but there are two main differences. (1) Spectral+post refines the features by using top-n gallery items of ranking list, while our method augments a memory in the network. (2) Post processing in spectral+post is a non-parametric operation, and our method is a parametric model and can be trained in an end-to-end fashion.

\noindent \textbf{Few-shot Learning.}  We compare our method and state-of-the-arts on $\emph{mini}$ImageNet and $\emph{tiered}$ImageNet, and the results are reported in Tab.\ref{tab:sota_on_DukeMTMC} and Tab.\ref{tab:sota_on_cuhk03} in terms of 
mean accuracy with 95\% confidence interval. It can be seen that the transductive few-shot learning methods outperform most of original few-shot learning methods, since transductive few-shot methods are allowed to explore unlabelled test samples in the test stage.  Meanwhile, our approach significantly outperforms the compared methods, especially in the one-shot scenario. It achieves 60.20 ${\pm0.23}$\% and 60.04${\pm 0.28}$\%  1-shot accuracies on \emph{mini}ImageNet and \emph{tired}ImageNet, respectively. Notably, the 5-shot accuracy of FEAT\cite{liu2019fewTPN} is 0.49\% better than ours in transductive setting on \emph{mini}ImageNet. One possible reason is that Feat employs self-attention mechanism to select related instances to enhance the feature embeddings, which is similar to our attention-based aggregation. 

\section{Conclusion}
In this work, we have proposed a novel Memory-based Neighbourhood Embedding (MNE) approach.  It enhances the feature embeddings of a single image instance by exploiting the information and relationships in the instance's neighbourhood.  Our approach augments the network with an episodic memory, which can provide the features of more relevant neighbours in training and testing. The neighbours are organized as a tree, and their features are gradually aggregated to the target instance in a bottom-up manner. Besides, the feature aggregation is based on a supervised attention strategy. We carefully verified the effectiveness of various components in MNE on image search and few-shot learning tasks, and our method can achieve the state-of-the-art performances on both tasks.
\vspace{-1em}

\section*{Acknowledgement}
\vspace{-1em}
This work is supported by the National Natural Science Foundation of China (Grant No. 61371192), the Key Laboratory Foundation of the Chinese Academy of Sciences (CXJJ-17S044) and the Fundamental Research Funds for the Central Universities (WK2100330002, WK3480000005).

{\small
\bibliographystyle{ieee_fullname}
\bibliography{mre}
}

\end{document}